\documentclass[10pt,twocolumn,letterpaper]{article}
\pdfoutput=1
\usepackage{cvpr}
\usepackage{times}
\usepackage{epsfig}
\usepackage{graphicx}
\usepackage{amsmath}
\usepackage{amssymb}

\usepackage{algorithm}
\usepackage{algorithmic}
\usepackage{multirow}

\usepackage[pagebackref=true,breaklinks=true,letterpaper=true,colorlinks,bookmarks=false]{hyperref}

\hypersetup{
  pdfinfo={
    Title={Learning Not to Learn},
    Author={Byungju Kim}
  }
}
\usepackage{pdfpages}

\cvprfinalcopy 

\def\cvprPaperID{4452} 

\ifcvprfinal\fi
\begin{document}
\title{Learning Not to Learn: Training Deep Neural Networks with Biased Data}
\author{Byungju Kim$^1$ \qquad  Hyunwoo Kim$^2$ \qquad Kyungsu Kim$^3$ \qquad Sungjin Kim$^3$ \qquad Junmo Kim$^1$\\
School of Electrical Engineering, KAIST, South Korea$^1$\\
Beijing Institute of Technology$^2$\\
Samsung Research$^3$\\
{\tt\small \{byungju.kim,junmo.kim\}@kaist.ac.kr}\\
{\tt\small \{hwkim\}@bit.edu.cn}\\
{\tt\small \{ks0326.kim,sj9373.kim\}@samsung.com}
\vspace*{-5mm}
}
\maketitle

\begin{abstract}
We propose a novel regularization algorithm to train deep neural networks, in which data at training time is severely biased.
Since a neural network efficiently learns data distribution, a network is likely to learn the bias information to categorize input data.
It leads to poor performance at test time, if the bias is, in fact, irrelevant to the categorization.
In this paper, we formulate a regularization loss based on mutual information between feature embedding and bias.
Based on the idea of minimizing this mutual information, we propose an iterative algorithm to unlearn the bias information.
We employ an additional network to predict the bias distribution and train the network adversarially against the feature embedding network.
At the end of learning, the bias prediction network is not able to predict the bias not because it is poorly trained, but because the feature embedding network successfully unlearns the bias information.
We also demonstrate quantitative and qualitative experimental results which show that our algorithm effectively removes the bias information from feature embedding.
\end{abstract}

\section{Introduction}
Machine learning algorithms and artificial intelligence have been used in wide ranging fields.
The growing variety of applications has resulted in great demand for robust algorithms. 
The most ideal way to robustly train a neural network is to use suitable data free of bias.
However great effort is often required to collect well-distributed data.
Moreover, there is a  lack of consensus as to what constitutes well-distributed data.

Apart from the philosophical problem, the data distribution significantly affects the characteristics of networks, as current deep learning-based algorithms learn directly from the input data.
If biased data is provided during training, the machine perceives the biased distribution as meaningful information.
This perception is crucial because it weakens the robustness of the algorithm and unjust discrimination can be introduced.

\begin{figure}[t]
\begin{center}
\includegraphics[width=1.\linewidth]{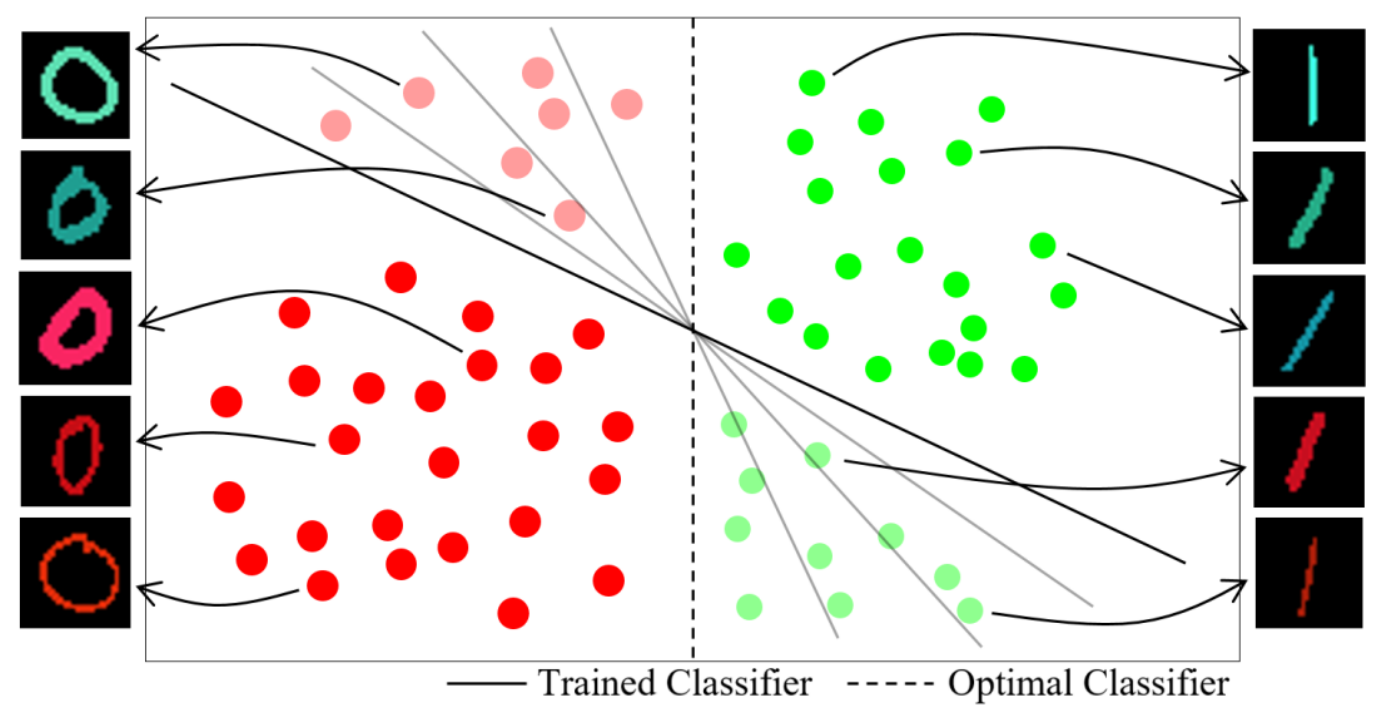}
\end{center}
\vspace*{-2mm}
\caption{Detrimental effect of biased data. The points colored with high saturation indicate samples provided during training, while the points with low saturation would appear in test scenario. Although the classifier is well-trained to categorize the training data, it performs poorly with test samples because the classifier learns the latent bias in the training samples.}
\vspace*{-4mm}
\label{fig:teaser}
\end{figure}

A similar concept has been explored in the literature and is referred to as \textit{unknowns} \cite{attenberg2015beat}.
The authors categorized unknowns as follows: \textit{known unknowns} and \textit{unknown unknowns}.
The key criterion differentiating these categories is the confidence of the predictions made by the trained models.
The unknown unknowns correspond to data points that the model's predictions are wrong with high confidence, e.g. high softmax score, whereas the known unknowns represent mispredicted data points with low confidence.
Known unknowns have better chance to be detected as the classifier's confidence is low, whereas unknown unknowns are much difficult to detect as the classifier generates high confidence score.

In this study, the data bias we consider has a similar flavor to the unknown unknowns in \cite{attenberg2015beat}.
However, unlike the unknown unknowns in \cite{attenberg2015beat}, the bias does not represent data points themselves.
Instead, bias represents some attributes of data points, such as color, race, or gender.

Figure~\ref{fig:teaser} conceptually shows how biased data can affect an algorithm.
The horizontal axis represents shape space of the digits, while the vertical axis represents color space, which is biased information for digit categorization.
In practice, shape and color are independent features, so a data point can appear anywhere in Figure~\ref{fig:teaser}.
However, let us assume that only the data points with high saturation are provided during training, but the points with low saturation are present in the test scenario (yet are not accessible during the training).
If a machine learns to categorize the digits, each solid line is a proper choice for the decision boundary.
Every decision boundary categorizes the training data perfectly, but it performs poorly on the points with low saturation.
Without additional information, learning of the decision boundary is an ill-posed problem, multiple decision boundaries can be determined that perfectly categorize the training data.
Moreover, it is likely that a machine would utilize the color feature because it is a simple feature to extract.

To fit the decision boundary to the optimal classifier in Figure~\ref{fig:teaser}, we require simple prior information: \textit{Do not learn from color distribution.}
To this end, we propose a novel regularization loss, based on mutual information, to train deep neural networks, which prevents learning of a given bias.
In other words, we regulate a network to minimize the mutual information shared between the extracted feature and the bias we want to unlearn.
Hereafter, the bias that we intend to unlearn is referred to the \textit{target bias}.
For example, the target bias is the color in Figure~\ref{fig:teaser}.
Prior to the unlearning of target bias, we assume that the existence of data bias is known and that the relevant meta-data, such as statistics or additional labels corresponding to the semantics of the biases are accessible.
Then, the problem can be formulated in terms of an adversarial problem.
In this scenario, one network has been trained to predict the target bias.
The other network has been trained to predict the label, which is the main objective of the network, while minimizing the mutual information between the embedded feature and the target bias.
Through this adversarial training process, the network can learn how to predict labels independent of the target bias.

Our main contributions can be summarized as follows:
Firstly, we propose a novel regularization term, based on mutual information, to unlearn target bias from the given data.
Secondly, we experimentally show that the proposed regularization term minimizes the detrimental effects of bias in the data.
By removing information relating to the target bias from feature embedding, the network was able to learn more informative features for classification.
In all experiments, networks trained with the proposed regularization loss showed performance improvements.
Moreover, they achieved the best performance in the most experiments.
Lastly, we propose bias planting protocols for public datasets.
To evaluate bias removal problem, we intentionally planted bias to training set while maintaining test set unbiased.

\section{Related Works}
The existence of unknown unknowns was experimentally demonstrated by Attenberg \etal  in \cite{attenberg2015beat}.
The authors separated the decisions rendered by predictive models into four conceptual categories: known knowns, known unknowns, unknown knowns, and unknown unknowns.
Subsequently, the authors developed and participated in a ``\textit{beat the machine} challenge'', which challenged the participants to manually find the unknown unknowns to fool the machine.

Several approaches for identifying unknown unknowns have been also proposed~\cite{aaai_unks, bansal2018coverage}.
Lakkaraju \etal \cite{aaai_unks} proposed an automatic algorithm using the explore-exploit strategy.
Bansal and Weld proposed a coverage-based utility model that evaluates the coverage of discovered unknown unknowns~\cite{bansal2018coverage}.
These approaches rely on an oracle for a subset of test queries.
Rather than relying on an oracle, Alvi \etal~\cite{blindeye} proposed joint learning and unlearning method to remove bias from neural network embedding.
To \textit{unlearn} the bias, the authors applied confusion loss, which is computed by calculating the cross-entropy between classifier output and a uniform distribution.
Similar approaches, making networks to be confused, have been applied on various applications ~\cite{dann,caption_eccv2018,confusion_eccv2018}.

As mentioned by Alvi \etal in the paper~\cite{blindeye}, the unsupervised domain adaptation (UDA) problem is closely related to the biased data problem.
The UDA problem involves generalizing the network embedding over different domains~\cite{dann, domain_adaptation1, domain_adaptation2}.
The main difference between our problem and the UDA problem is that our problem does not assume the access to the target images and instead, we are aware of the description of the target bias.

Embracing the UDA problem, disentangling feature representation has been widely researched in the literature.
The application of disentangled features has been explored in detail~\cite{rotate_face,pose_face_recog}.
Using generative adversarial network~\cite{gan}, more methods to learn disentangled representation~\cite{infogan,mtan,disentangle_gan} have been proposed.
In particular, Chen \etal proposed the InfoGAN~\cite{infogan} method, which learns and preserves semantic context without supervision.

These studies highlighted the importance of feature disentanglement, which is the first step in understanding the information contained within the feature.
Inspired by various applications, we have attempted to remove certain information from the feature.
In contrast to the InfoGan~\cite{infogan}, we minimize the mutual information in order \textit{not} to learn.
However, removal of information is an antithetical concept to learning and is also referred to as \textit{unlearning}.
Although the concept itself is the complete opposite of learning, it can help learning algorithms.
Herein, we describe an algorithm for removing target information and present experimental results and analysis to support the proposed algorithm.

\section{Problem Statement}
In this section, we formulate a novel regularization loss, which minimizes the undesirable effects of biased data, and describe the training procedure.
The notations should be defined prior to introduction of the formulation.
Unless specifically mentioned, all notation refers to the following terms hereafter.
Assume we have an image $x \in \mathcal{X}$ and corresponding label $y_{x}\in\mathcal{Y}$.
We define a set of bias, $\mathcal{B}$, which contains every possible target bias that $\mathcal{X}$ can possess.
In Figure~\ref{fig:teaser}, $\mathcal{B}$ is a set of possible colors, while $\mathcal{Y}$ represents a set of digit classes.
We also define a latent function $b : \mathcal{X} \rightarrow \mathcal{B}$, where $b(x)$ denotes the target bias of $x$.
We define random variables $X$ and $Y$ that have the value of $x$ and $y_{x}$ respectively.

\begin{figure}[t]
\begin{center}
\includegraphics[width=0.9\linewidth]{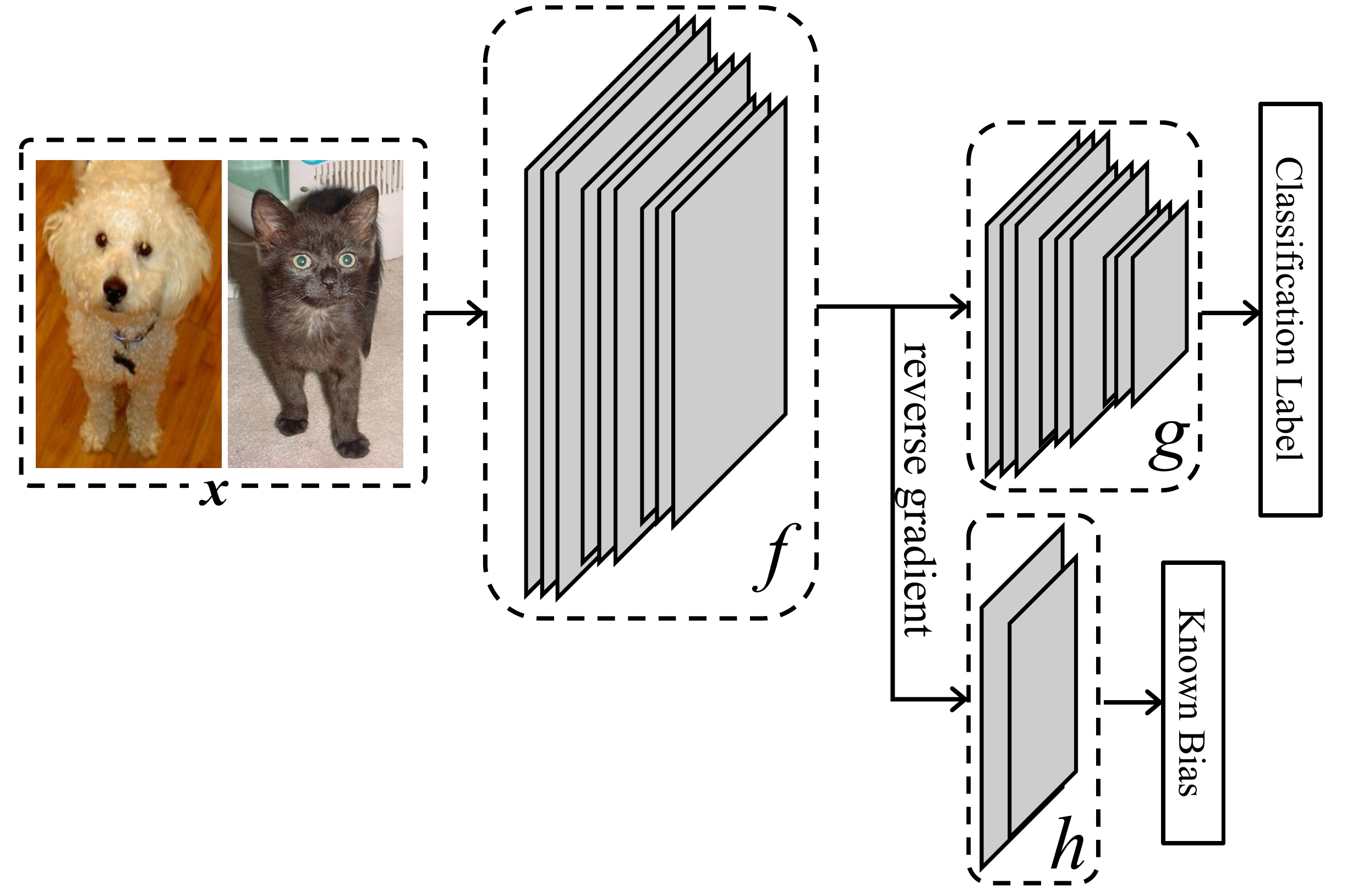}
\end{center}
\vspace*{-2mm}
\caption{Overall architecture of deep neural network. The network $g \circ f$ is implemented with ResNet-18~\cite{resnet} for real images and plain network with four convolution layers for MNIST images.}
\label{fig:arch}
\end{figure}

The input image $x$ is fed into the feature extraction network $ f : \mathcal{X} \rightarrow \mathbb{R}^{K}$, where $K$ is the dimension of the feature embedded by $f$.
Subsequently, the extracted feature, $f(x)$, is fed forward through both the label prediction network $ g : \mathbb{R}^{K} \rightarrow \mathcal{Y}$, and bias prediction network $ h : \mathbb{R}^{K} \rightarrow \mathcal{B}$.
The parameters of each network are defined as $\theta_{f}, \theta_{g}$, and $\theta_{h}$ with the subscripts indicating their specific network.
Figure~\ref{fig:arch} describes the overall architecture of the neural networks.
However, we do not explicitly designate a detailed architecture, since our regularization loss is applicable to arbitrary network architectures.

\subsection{Formulation}
The objective of our work is to train a network that performs robustly with unbiased data during test time, even though the network is trained with biased data.
The data bias has following characteristic:
\begin{equation}
\label{eq:mutual_info_b_Y}
    \mathcal{I}(b(X^{train});Y) \gg \mathcal{I}(b(X^{test});Y) \approx 0,
\end{equation}
where $X^{train}$ and $X^{test}$ denote the random variable sampled during the training and test procedure, respectively, and $\mathcal{I}(\cdot;\cdot)$ denotes the mutual information.
Biased training data results in the biased networks, so that the network relies heavily on the bias of the data:
\begin{equation}
\label{eq:mutual_info_b_feat}
    \mathcal{I}(b(X);g(f(X))) \gg 0.
\end{equation}
To this end, we add the mutual information to the objective function for training networks.
We minimize the mutual information over $f(X)$, instead of $g(f(X))$.
It is adequate because the label prediction network, $g$, takes $f(X)$ as its input.
From a standpoint of $g$, the training data is not biased if the network $f$ extracts no information of the target bias.
In other words, extracted feature $f(x)$ should contain no information of the target bias, $b(x)$.
Therefore, the training procedure is to optimize the following problem:
\begin{equation}
\label{eq:initial_form_obj}
    \min_{\theta_f,\theta_g} \ \mathbb{E}_{\tilde{x}\sim P_{X}(\cdot)}[\mathcal{L}_{c}(y_{\tilde{x}},g(f(\tilde{x})))] + \lambda \mathcal{I}(b(X);f(X)),
\end{equation}
where $\mathcal{L}_{c}(\cdot,\cdot)$ represents the cross-entropy loss, and $\lambda$ is a hyper-parameter to balance the terms.

The mutual information in Eq.~(\ref{eq:initial_form_obj}) can be equivalently expressed as follows: 
\begin{equation}
\label{eq:mutual_information_b_f_reform}
  \mathcal{I}(b(X);f(X)) = H(b(X)) - H(b(X) | f(X)),
\end{equation}
where $H(\cdot)$ and $H(\cdot|\cdot)$ denote the marginal and conditional entropy, respectively.
Since the marginal entropy of bias is constant that does not depend on $\theta_f$ and $\theta_g$, $H(b(X))$ can be omitted from the optimization problem, and we try to minimize the negative entropy, $-H(b(X)|f(X))$.
%
%
%
%
%
%
Eq.~(\ref{eq:mutual_information_b_f_reform}) is difficult to directly minimize as it requires the posterior distribution, $P(b(X)|f(X))$.
Since it is not tractable in practice, minimizing the Eq.~(\ref{eq:mutual_information_b_f_reform}) is reformulated using an auxiliary distribution, $Q$, with an additional equality constraint:
\begin{equation}
\label{eq:mi_reform_eq_const}
\begin{split}
    \min_{\theta_f}&\  \mathbb{E}_{\tilde{x}\sim P_{X}(\cdot)}[\mathbb{E}_{\tilde{b}\sim Q(\cdot|f(\tilde{x}))}[\log Q(\tilde{b}|f(\tilde{x}))]] \\
     s.t.& \ \ Q(b(X)|f(X))=P(b(X)|f(X)).
\end{split}
\end{equation}
The benefit of using the distribution $Q$ is that we can directly calculate the objective function.
Therefore, we can train the feature extraction network, $f$, under the equality constraint.

\subsection{Training Procedure}
As the equality constraint in Eq.~(\ref{eq:mi_reform_eq_const}) is difficult to meet (especially in the beginning of the training process), we modify the equality constraint into minimizing KL divergence between $P$ and $Q$, so that $Q$ gets closer to $P$ as learning progresses.
We relax the Eq.~(\ref{eq:mi_reform_eq_const}), so that the auxiliary distribution, $Q$, could be used to approximate the posterior distribution.
The relaxed regularization loss, $\mathcal{L}_{MI}$, is as follows:
\begin{equation}
\label{eq:loss_lagrange}
\begin{split}
    \mathcal{L}_{MI}& = \ \mathbb{E}_{\tilde{x}\sim P_{X}(\cdot)}[\mathbb{E}_{\tilde{b}\sim Q(\cdot|f(\tilde{x}))}[\log Q(\tilde{b}|f(\tilde{x}))]] \\
    &+ \mu D_{KL}(P(b(X)|f(X)) || Q(b(X)|f(X))),
\end{split}
\end{equation}
where $D_{KL}$ denotes the KL-divergence and $\mu$ is hyper-parameter which balances the two terms.
Similar to the method proposed by Chen \etal \cite{infogan}, we parametrize the auxiliary distribution, $Q$, as the bias prediction network, $h$.
Note that we will train network $h$, so that the KL-divergence is minimized.
Provided that the distribution $Q$ implemented by network $h$ converges to $P(b(X)|f(X))$, we only need to train network $f$ so that the first term in Eq.~(\ref{eq:loss_lagrange}) is minimized.

Although the posterior distribution, $P(b(X)|f(X))$, is not tractable, the bias prediction network, $h$, is expected to be trained to stochastically approximate $P(b(X)|f(X))$, if we train the network with $b(X)$ as the label with SGD optimizer.
Therefore, we relax the KL-divergence of Eq.~(\ref{eq:loss_lagrange}) with expectation of the cross-entropy loss between $b(X)$ and $h(f(X))$, and we train network $h$ so that bias prediction loss, $\mathcal{L}_{\mathcal{B}}$, is minimized.
\begin{equation}
\label{eq:loss_mi_complete}
    \mathcal{L}_{\mathcal{B}}(\theta_f,\theta_h) = \mathbb{E}_{\tilde{x}\sim P_{X}(\cdot)}[\mathcal{L}_{c}(b(\tilde{x}),h(f(\tilde{x})))].
\end{equation}
Although training network $h$ alone to minimize Eq.~(\ref{eq:loss_mi_complete}) is enough to make $Q$ closer to $P$, it will be additionally beneficial to train $f$ to maximize Eq.~(\ref{eq:loss_mi_complete}) in an adversarial way, i.e. to let the networks $f$ and $h$ play the minimax game.
The intuition is that the feature extracted by network $f$ is making the bias prediction difficult.
As $f$ is trained to minimize the first term in Eq.~(\ref{eq:loss_lagrange}), we can reformulate Eq.~(\ref{eq:loss_lagrange}) using  $\mathcal{L}_{\mathcal{B}}$ instead of KL-divergence as follows:

\begin{equation}
\label{eq:minmax_game}
\begin{split}
    \min_{\theta_f} \max_{\theta_h} \ \ \mathbb{E}_{\tilde{x}\sim P_{X}(\cdot)}[&\mathbb{E}_{\tilde{b}\sim Q(\cdot|f(\tilde{x}))}[\log Q(\tilde{b}|f(\tilde{x}))]] \\
     -\ & \mu \mathcal{L}_{\mathcal{B}}(\theta_f,\theta_h)
\end{split}
\end{equation}

We train $h$ to correctly predict the bias, $b(X)$, from its feature embedding, $f(X)$.
We train $f$ to minimize the negative conditional entropy.
The network $h$ is fixed while minimizing the negative conditional entropy.
The network $f$ is also trained to maximize the cross-entropy to restrain $h$ from predicting $b(X)$.
Together with the primal classification problem, the minimax game is formulated as follows:
\begin{equation}
\label{eq:minimax_obj}
\begin{split}
    \min_{\theta_f,\theta_g} \max_{\theta_h} \ \ &\mathbb{E}_{\tilde{x}\sim P_{X}(\cdot)}[\mathcal{L}_{c}(y_{\tilde{x}},g(f(\tilde{x}))) \\
    + \ &\lambda\mathbb{E}_{\tilde{b}\sim Q(\cdot|f(\tilde{x}))}[\log Q(\tilde{b}|f(\tilde{x}))]] \\
    - \ &\mu\mathcal{L}_{\mathcal{B}}(\theta_f,\theta_h)
\end{split}
\end{equation}

In practice, the deep neural networks, $f$, $g$ and $h$, are trained with both adversarial strategy \cite{gan, infogan} and gradient reversal technique \cite{dann}.
Early in learning, $g \circ f$ are rapidly trained to classify the label using the bias information.
Then $h$ learns to predict the bias, and $f$ begins to learn how to extract feature embedding independent of the bias.
At the end of the training, $h$ regresses to the poor performing network not because the bias prediction network,  $h$, diverges, but because $f$ unlearns the bias, so the feature embedding, $f(X)$, does not have enough information to predict the target bias.

\begin{figure*}[t]
\begin{center}
   \includegraphics[width=1.\linewidth]{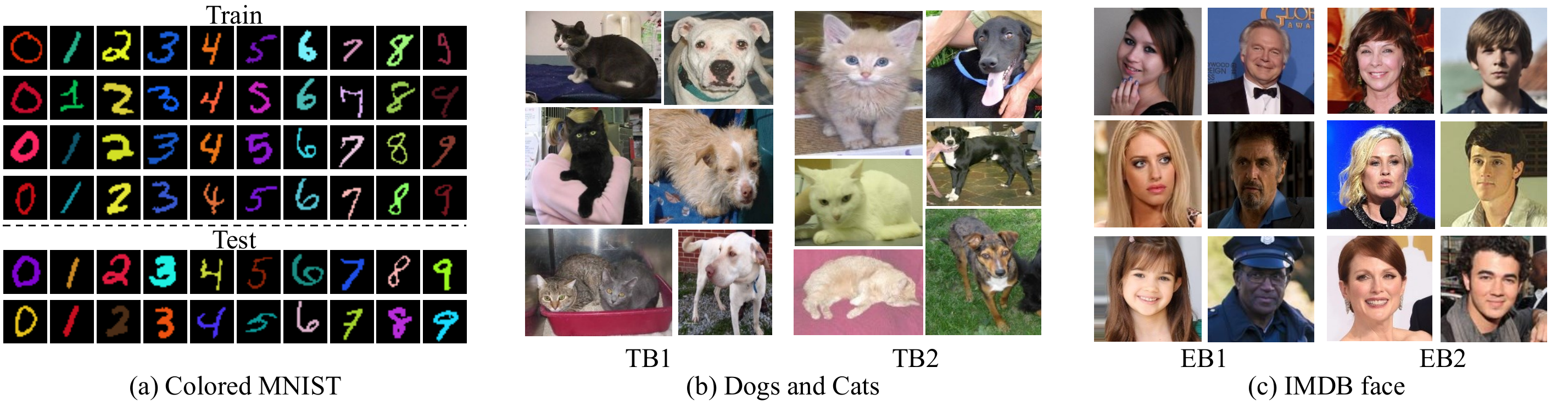}
\end{center}
\vspace*{-2mm}
\caption{Examples of datasets with intentionally planted bias. (a) We modified the MNIST data~\cite{mnist} to plant color bias to training images. A mean color has been designated for each class, so a classifier can easily predict the digit with color. (b) TB1 is a set of bright dogs and dark cats, whereas TB2 contains dark dogs and bright cats. Similar to the colored MNIST, a classifier can predict whether an image is dog or cat with its color. (c) IMDB face dataset contains age and gender labels. EB1 and EB2 differ on the correlation between age and gender. Predicting age enables an algorithm to predict gender.}
\vspace*{-1mm}
\label{fig:dataset_example}
\end{figure*}

\section{Dataset}
Most existing benchmarks are designed to evaluate a specific problem.
The collectors often split the dataset into train/test sets exquisitely.
However, their efforts to maintain the train/test split to obtain an identical distribution obscures our experiment.
Thus, we intentionally planted bias to well-balanced public benchmarks to determine whether our algorithm could unlearn the bias.

\subsection{Colored MNIST}
We planted a color bias into the MNIST dataset~\cite{mnist}.
To synthesize the color bias, we selected ten distinct colors and assigned them to each digit category as their mean color.
Then, for each training image, we randomly sampled a color from the normal distribution of the corresponding mean color and provided variance, and colorized the digit.
Since the variance of the normal distribution is a parameter that can be controlled, the amount of the color bias in the data can be adjusted.
For each test image, we randomly choose a mean color among the ten pre-defined colors and followed the same colorization protocol as for the training images.
Each sub-datasets are denoted as follows:
\begin{itemize}
    \item Train-$\sigma^2$: Train images with colors sampled with $\sigma^2$
    \item Test-$\sigma^2$: Test images with colors sampled with $\sigma^2$
\end{itemize}
Since the digits in the test sets are colored with random mean colors, the Test-$\sigma^2$ sets are unbiased.
We varied $\sigma^2$ from 0.02 to 0.05 with a 0.005 interval.
Smaller values of $\sigma^2$ indicate more bias in the set.
Thus, Train-0.02 is the most biased set, whereas Train-0.05 is the least biased.

Figure~\ref{fig:dataset_example} (a) shows samples from the colored MNIST, where the images in the training set show that the color and digit class are highly correlated.
The color of the digit contains sufficient information to categorize the digits in the training set, but it is insufficient for the images in the test set.
Recognizing the color would rather disrupt the digit categorization.
Therefore, the color information must be removed from the feature embedding.

\subsection{Dogs and Cats} 
We evaluated our algorithm with the dogs and cats database, developed by \textit{kaggle}~\cite{kaggle}.
The original database is a set of 25K images of dogs and cats for training and 12,500 images for testing.
Similar to \cite{aaai_unks}, we manually categorized the data according to the color of the animal: bright, dark, and other.
Subsequently, we split the images into three subsets.
\begin{itemize}
    \item Train-biased 1 (TB1) : bright dogs and dark cats.
    \item Train-biased 2 (TB2) : dark dogs and bright cats.
    \item Test set: All 12,500 images from the original test set.
\end{itemize}
The images categorized as \textit{other} are images featuring white cats with dark brown stripes or dalmatians.
They were not used in our training sets due to their ambiguity.
In turn, TB1 and TB2 contain 10,047 and 6,738 images respectively.
The constructed dogs and cats dataset is shown in Figure~\ref{fig:dataset_example} (b), with each set containing a color bias.
For this dataset, the bias set $\mathcal{B} = \{ \mbox{dark, bright}\}$.
Unlike TB1 and TB2, the test set does not contain color bias.

On the other hand, the ground truth labels for test images are not accessible, as the data is originally for competition~\cite{kaggle}.
Therefore, we trained an oracle network (ResNet-18~\cite{resnet}) with all 25K training images.
For the test set, we measured the performance based on the result from the oracle network.
We presumed that the oracle network could accurately predict the label.

\subsection{IMDB Face}
The IMDB face dataset~\cite{IMDB} is a publicly available face image dataset.
It contains 460,723 face images from 20,284 celebrities along with information regarding their age and gender.
Each image in the IMDB face dataset is a cropped facial image.
As mentioned in \cite{IMDB,blindeye}, the provided label contains significant noise.
To filter out misannotated images, we used pretrained networks~\cite{age_gender_cls} on Adience benchmark~\cite{other_dataset} designed for age and gender classification.
Using the pretrained networks, we estimated the age and gender for all the individuals shown in the images in the IMDB face dataset.
We then collected images where the both age and gender labels match with the estimation.
From this, we obtained a cleaned dataset with 112,340 face images, and the detailed cleaning procedure is described in the supplementary material.

\begin{figure}[t]
\begin{center}
   \includegraphics[width=1.\linewidth]{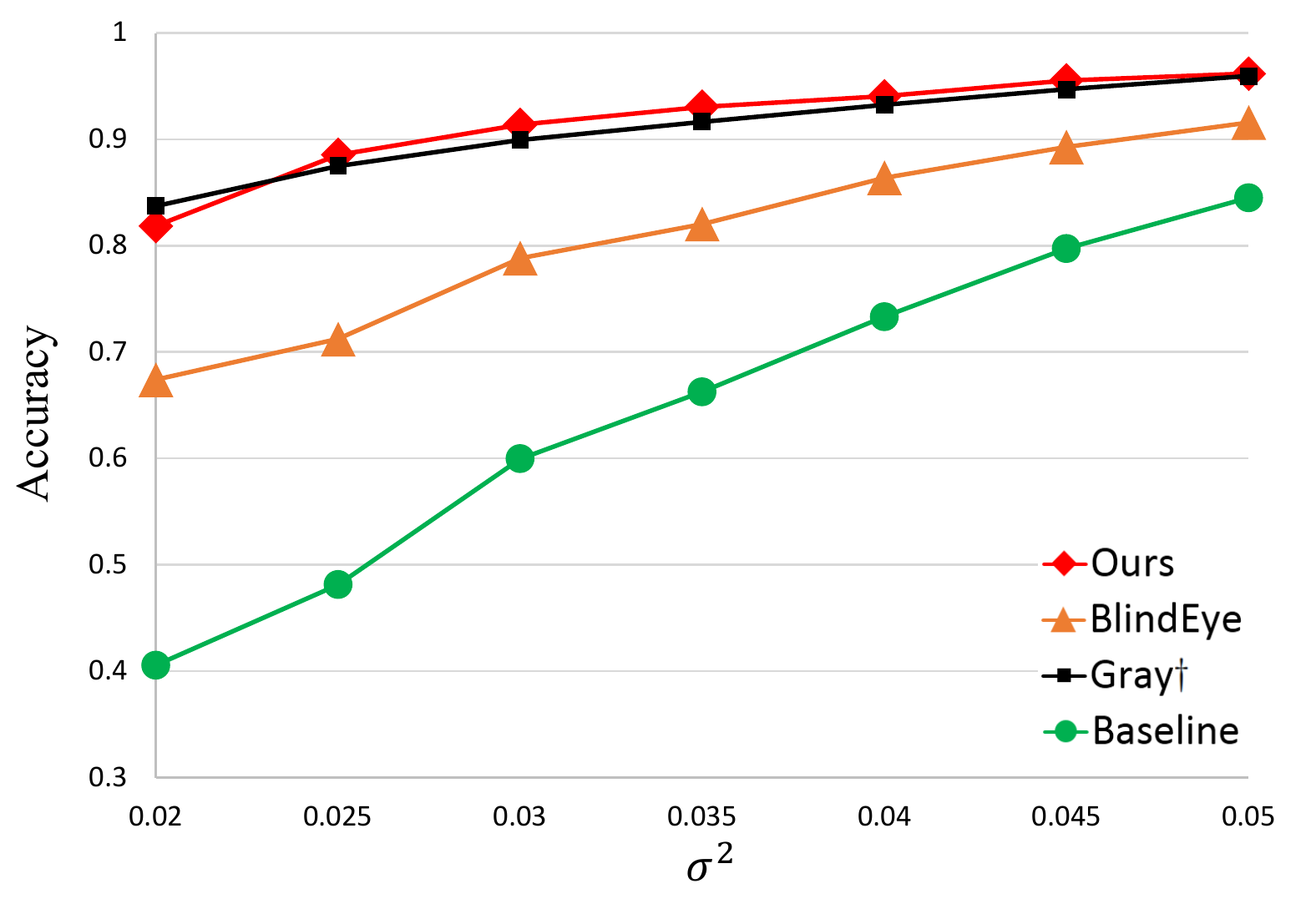}
\end{center}
\vspace*{-4mm}
\caption{Evaluation results on colored MNIST dataset. $\dagger$ denotes that it is evaluated with grayscale-converted images. The model denoted as Gray was trained with images converted into grayscale; it is trained with significantly mitigated bias. Compared to the baseline and BlindEye algorithm~\cite{blindeye}, our model shows outperforming results. Note that our result shows comparable performance with grayscale model. It implies that the network was successfully trained to extract feature embedding a lot more independent of the bias.}
\vspace*{-2mm}
\label{fig:mnist_result}
\end{figure}

\begin{figure*}[t]
\begin{center}
   \includegraphics[width=1.\linewidth]{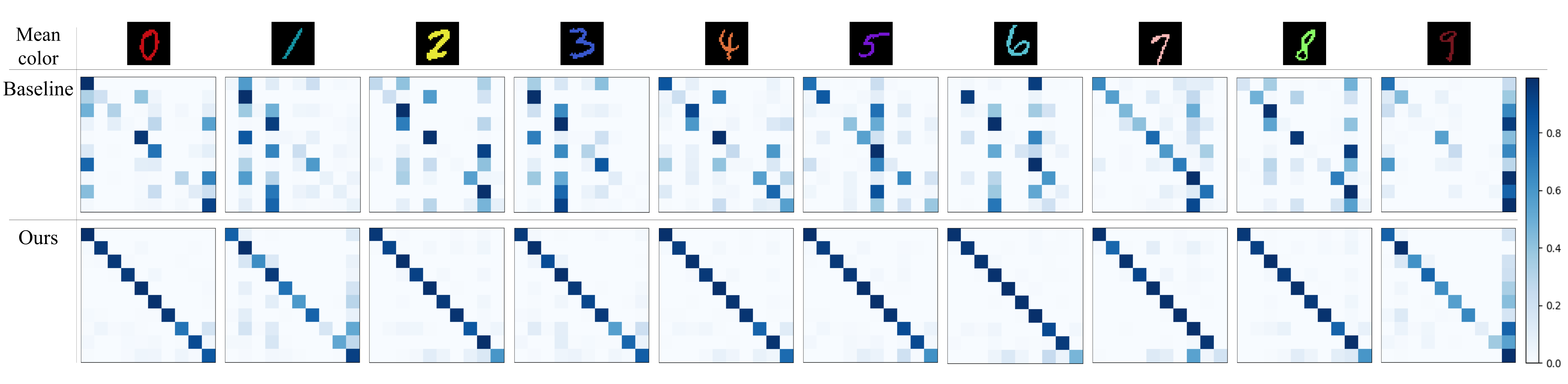}
\end{center}
\vspace*{-5mm}
\caption{Confusion matrices with test images colored by single mean color. 
Top row denotes the mean colors and their corresponding digit classes in training data. 
The confusion matrices of baseline model show the network is biased owing to the biased data.
On the contrary, the networks trained by our algorithm are not biased to the color although they were trained with the same training data with the baseline.}
\label{fig:confusion_matrix}
\end{figure*}

Similar to the protocol from \cite{blindeye}, we classified the cleaned IMDB images into three biased subsets.
We first withheld 20\% of the cleaned IMDB images as the test set, then split the rest of the images as follows:
\begin{itemize}
\small
    \item Extreme bias 1 (EB1): women aged 0-29, men aged 40+
    \item Extreme bias 2 (EB2): women aged 40+, men aged 0-29
    \item Test set: 20\% of the cleaned images aged 0-29 or 40+
\end{itemize}
\normalsize
As a result, EB1 and EB2 contain 36,004 and 16,800 facial images respectively, and the test set contains 13129 images.
Figure~\ref{fig:dataset_example} (c) shows that both EB1 and EB2 are biased with respect to the age.
Although it is not as clear as the color bias in Figure~\ref{fig:dataset_example} (a) and (b), 
EB1 consists of younger female and older male celebrities, whereas EB2 consists of younger male and older female celebrities. When gender is target bias, $\mathcal{B} = \{ \mbox{male, female}\}$, and when age is target bias, $\mathcal{B}$ is their age.

\section{Experiments}
\subsection{Implementation}
In the following experiments, we removed three types of target bias: color, age, and gender.
The age and gender labels were provided in IMDB face dataset, therefore $\mathcal{L}_{\mathcal{B}}(\theta_f,\theta_{h})$ was optimized with supervision.
On the other hand, the color bias was removed via self-supervision.
To construct color labels, we first sub-sampled the images by factor of 4.
In addition, the dynamic range of color, 0-255, was quantized into eight even levels.

For the network architecture, we used ResNet-18 \cite{resnet} for real images and plain network with four convolution layers for the colored MNIST experiments.
The network architectures correspond to the parametrization of $g\circ f$.
In the case we used ResNet-18, $g$ was implemented as two residual blocks on the top, while $f$ represents the rest.
For plain network for colored MNIST, both $g$ and $f$ consist of two convolution layers.
ResNet-18 was pretrained with Imagenet data~\cite{ILSVRC15} except for the last fully connected layer.
We implemented $h$ with two convolution layers for color bias and single fully connected layer for gender and age bias.
Every convolution layer is followed by batch normalization \cite{batchnorm} and ReLU activation layers.
All the evaluation results were averaged to be presented in this paper.


\subsection{Results}
We compare our training algorithm with other methods that can be used for this task.
The performance of the algorithms mentioned in this section were re-implemented based on the literature.

\noindent
\textbf{Colored MNIST.}
The amount of bias in the data was controlled by adjusting the value of $\sigma^2$.
A network was trained for each $\sigma^2$ value from 0.02 to 0.05 and was evaluated with the corresponding test set with the same $\sigma^2$.
Since a color for each image was sampled with a given $\sigma^2$, smaller $\sigma^2$ implies severer color bias.
Figure~\ref{fig:mnist_result} shows the evaluation results of the colored MNIST.
The baseline model represents a network trained without additional regularization and the baseline performance can roughly be used as an indication of training data bias.
The algorithm denoted as ``BlindEye'' represents a network trained with confusion loss~\cite{blindeye} instead of our regularization.
The other algorithm, denoted as ``Gray'', represents a network trained with grayscale images and it was also tested with grayscale images.
For the given color biased data, we converted the color digits into grayscale.
Conversion into grayscale is a trivial approach that can be used to mitigate the color bias.
We presume that the conversion into grayscale does not reduce the information significantly since the MNIST dataset was originally provided in grayscale.

The results of our proposed algorithm outperformed the BlindEye~\cite{blindeye} and baseline model with all values of $\sigma^2$.
Notably, we achieved similar performance as the model trained and tested with grayscale images.
Since we converted images in both training and test time, the network is much less biased.
In most experiments, our model performed slightly better than the gray algorithm, suggesting that our regulation algorithm can effectively remove the target bias and encourage a network to extract more informative features.

To analyze the effect of the bias and proposed algorithm, we re-colored the test images.
We sampled with the same protocol, but with fixed mean color, which was assigned to one of the ten digit classes of the biased training data.
Figure~\ref{fig:confusion_matrix} shows the confusion matrices drawn by the baseline and our models with the re-colored test images.
The digits illustrated in the top row denotes the mean colors and their corresponding digit class in training set.
For example, the first digit, red zero, signifies the confusion matrices below are drawn by test images colored reddish regardless of their true label.
It also stands for a fact that every digit of category \textit{zero} in training data is colored reddish.

In Figure~\ref{fig:confusion_matrix}, the matrices of the baseline show vertical patterns, some of which are shared, such as digits 1 and 3.
The mean color for class 1 is \textit{teal}; in RGB space it is (0, 128, 128).
The mean color for class 3 is similar to that of class 1.
In RGB space, it is (0, 149, 182) and is called \textit{bondi blue}.
This indicates that the baseline network is biased to the color of digit.
As observed from the figure, the confusion matrices drawn by our algorithm (bottom row) show that the color bias was removed.

\noindent
\textbf{Dogs and Cats.}
Table~\ref{table:cats_result} presents the evaluation results, where the baseline networks perform admirably, considering the complexity of the task due to the pretrained parameters.
As mentioned in \cite{shape_bias}, neural networks prefer to categorize images based on shape rather than color.
This encourages the baseline network to learn shapes, but the evaluation results presented in Table~\ref{table:cats_result} imply that the networks remain biased without regularization.

Similar to the experiment on the colored MNIST, simplest approach for reducing the color bias is to convert the images into grayscale.
Unlike the MNIST dataset, conversion would remove a significant amount of information.
Although the networks for grayscale images performed better than the baseline, Table~\ref{table:cats_result} shows that the networks remain biased to color.
This is likely because of the criterion that was used to implant the color bias.
Since the original dataset is categorized into bright and dark, the converted images contain a bias in terms of brightness.

We used gradient reversal layer (GRL)~\cite{dann} and adversarial training strategy~\cite{infogan, gan} as components of our optimization process.
To analyze the effect of each component, we ablated the GRL from our algorithm.
We also trained networks with both confusion loss~\cite{blindeye} and GRL, since they can be used in conjunction with each other.
Although the GRL was originally proposed to solve unsupervised domain adaptation problem~\cite{dann}, Table~\ref{table:cats_result} shows that it is beneficial for bias removal.
Together with either confusion loss or our regularization, we obtained the performance improvements.
Furthermore, GRL alone notably improved the performance suggesting that GRL itself is able to remove bias.

\begin{figure*}[t]
\begin{center}
   \includegraphics[width=1.\linewidth]{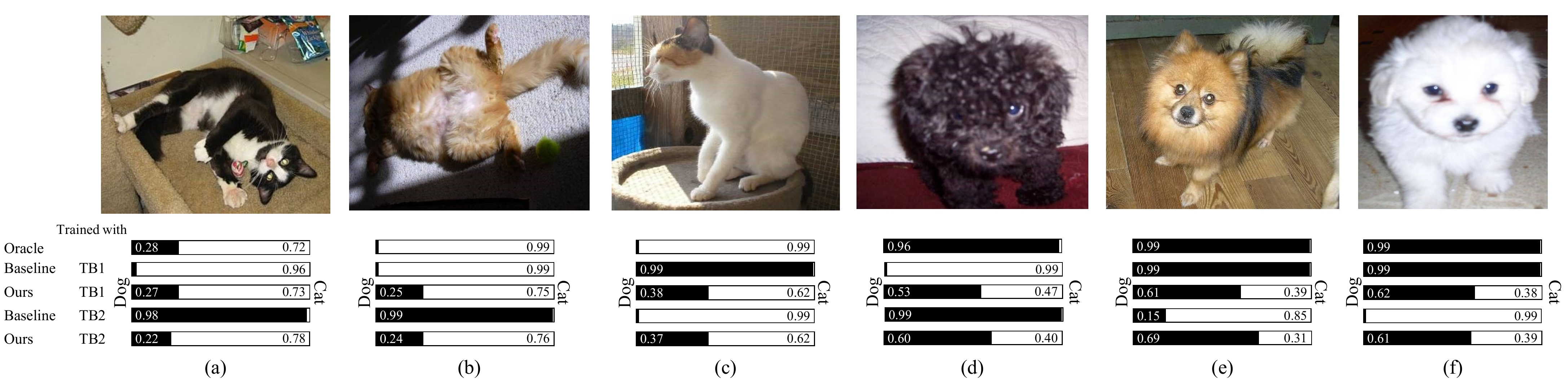}
\end{center}
\vspace*{-4mm}
\caption{Qualitative results of dogs and cats dataset.
The oracle model was trained with not only both TB1 and TB2, but also with images we categorized as \textit{other} color. 
For test images, prediction results of the oracle model were considered as their true labels. 
The stacked bar charts below the figures visualize prediction results by each model. 
The baseline models tend to predict depending on the color, whereas our model ignores the color information.}
\vspace*{-2mm}
\label{fig:cats_result}
\end{figure*}

Figure~\ref{fig:cats_result} shows the qualitative effect of our proposed regularization.
The prediction results of the baseline networks do not change significantly regardless of whether the query image is cat or dog if the colors are similar.
If a network is trained with TB1, the network predicts a dark image to be a cat and a bright image to be a dog.
If another network is trained with TB2, the network predicts a bright image to be a cat and a dark image to be a dog.
This implies that the baseline networks are biased to color.
On the other hand, networks trained with our proposed algorithm successfully classified the query images independent of their colors.
In particular, Figure~\ref{fig:cats_result} (c) and (f) were identically predicted by the baseline networks depending on their color.
After removing the color information from the feature embedding, the images were correctly categorized according to their appearance.

\begin{table}[t]
\begin{center}
\begin{tabular}{lccccc}
\hline
         & \multicolumn{2}{c}{Trained on TB1}  & & \multicolumn{2}{c}{Trained on TB2} \\ \cline{2-3}\cline{5-6}
Method   &    \:~TB2~\:     &       Test       & &     \:~TB1\:~    &     Test         \\ \hline\hline
Baseline &      .7498       &      .9254       & &       .6645      &     .8524        \\
Gray$\dagger$&  .8366       &      .9483       & &       .7192      &     .8687        \\
BlindEye~\cite{blindeye} &      .8525       &      .9517       & &       .7812      &     .9038        \\
GRL~\cite{dann}      &      .8356       &      .9462       & &       .7813      &     .9012        \\
BlindEye+GRL & \underline{.8937}&      .9582   & &       .8610      &     .9291        \\ \hline
Ours-adv &      .8853       & \underline{.9594}& &\underline{.8630} &\underline{.9298} \\
Ours     &  \bf{.9029}      &  \bf{.9638}      & &   \bf{.8726}     &  \bf{.9376}      \\
\hline
\end{tabular}
\end{center}
\vspace*{-1mm}
\caption{The evaluation results on dogs and cats dataset. 
All networks were evaluated with test set. 
Moreover, the networks trained with TB1 were additionally evaluated with TB2, and vice versa. 
$\dagger$ denotes that the network was tested with images converted into grayscale. 
The Ours-adv denotes a model trained with Eq.~(\ref{eq:minimax_obj}) without using gradient reversal layer.
The best performing result on each column is denoted as boldface and the second best result is underlined.}
\label{table:cats_result}
\end{table}

\begin{table}[t]
\begin{center}
\begin{tabular}{lccccc}
\hline
                        & \multicolumn{2}{c}{Trained on EB1} & & \multicolumn{2}{c}{Trained on EB2} \\ \cline{2-3}\cline{5-6} 
Method                  & ~~EB2~~            & Test          & & ~~EB1~~          & Test             \\ \hline
\multicolumn{1}{c}{}  & \multicolumn{5}{c}{Learn Gender,  Unlearn Age}                  \\ \cline{2-6} 
Baseline              &.5986             &.8442            & &.5784             &.6975             \\
BlindEye~\cite{blindeye}              &.6374             &.8556            & &.5733             &.6990             \\ 
Ours                  &\bf{.6800}        &\bf{.8666}       & &\bf{.6418}        &\bf{.7450}        \\
\hline
\multicolumn{1}{c}{}  & \multicolumn{5}{c}{Learn Age, Unlearn Gender}                  \\ \cline{2-6} 
Baseline              &.5430             &.7717            & &.4891             &.6197             \\
BlindEye ~\cite{blindeye}             &\bf{.6680}             &.7513            & &\bf{.6416}    &.6240             \\
Ours                  &.6527             &\bf{.7743}            & &.6218             &\bf{.6304}             \\ \hline
\end{tabular}
\end{center}
\vspace*{-1mm}
\caption{Evaluation results on IMDB face dataset. 
All networks were evaluated with test set and the other training set.
The best performing result on each column is denoted as boldface.}
\label{table:imdb_result}
\vspace*{-2mm}
\end{table}

\begin{figure*}[t]
\begin{center}
   \includegraphics[width=1.\linewidth]{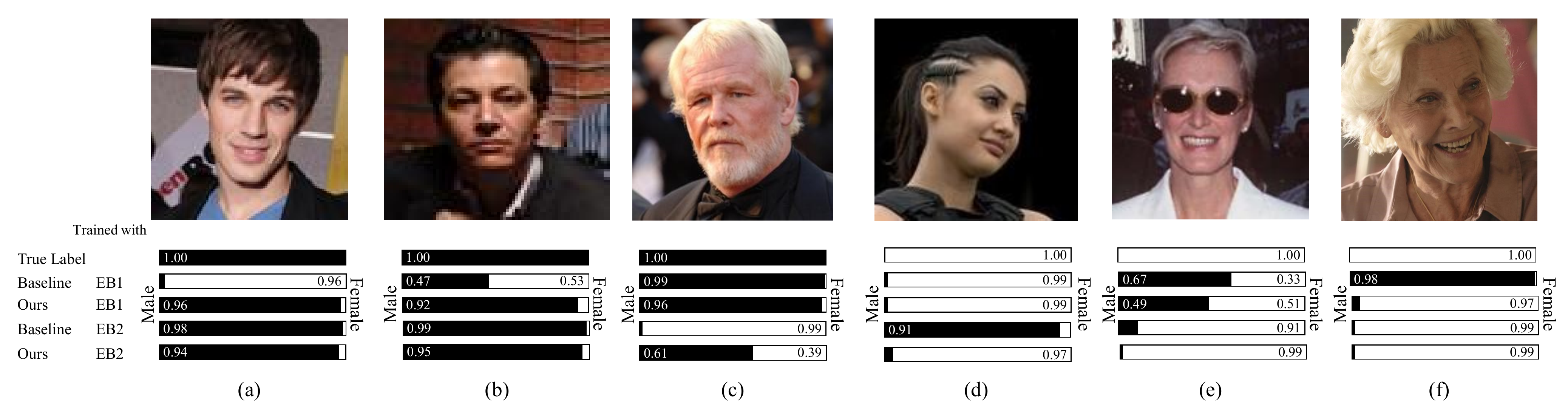}
\end{center}
\vspace*{-4mm}
\caption{Qualitative results of gender classification with IMDB face dataset. 
As in the Figure~\ref{fig:cats_result}, the stacked bar charts represent the prediction results. 
They show that the baseline models are biased to the age. On the other hand, the networks trained with proposed algorithm predict the gender independent of their age.}
\vspace*{-1mm}
\label{fig:imdb_result}
\end{figure*}


\noindent
\textbf{IMDB face.}
For the IMDB face dataset, we conducted two experiments; one to train the networks to classify age independent of gender, and one to train the networks to classify gender independent of age.
Table~\ref{table:imdb_result} shows the evaluation results from both experiments.
The networks were trained with either EB1 or EB2 and since they are extremely biased, the baseline networks are also biased.
By removing the target bias information from the feature embedding, overall performances are improved.
On the other hand, considering that gender classification is a two class problem, where random guessing achieves 50\% accuracy, the networks perform poorly on gender classification.
Although Table~\ref{table:imdb_result} shows that the performance improves after removing the target bias from the feature embedding, the performance improvement achieved using our algorithm is marginal compared to previous experiments with other datasets.
We presume that this is because of the correlation between age and gender.
In the case of color bias, the bias itself is completely independent of the categories.
In other words, an effort to unlearn the bias is purely beneficial for digit categorization.
Thus, removing color bias from feature embedding improved the performance significantly because the network is able to focus on learning shape features.
Unlike the color bias, age and gender are not completely independent features. 
Therefore, removing bias information from feature embedding would not be completely beneficial.
This suggests that a deep understanding of the specific data bias must precede the removal of bias.

Figure~\ref{fig:imdb_result} shows the qualitative effect of regularization on the gender classification task.
Young, mid-age, and old individuals with both male and female are presented.
Similar to Figure~\ref{fig:cats_result}, it implies that the baseline networks are biased toward age.
The baseline network trained with EB1 predicted both young male and young female images (Figure~\ref{fig:imdb_result} (a) and (d)) as female with high confidence.
Meanwhile, the network trained with EB2 predicted the same images as the exact opposite gender with high confidence.
Upon removal of age bias, the networks were trained to correctly predict the gender.

\section{Conclusion}
In this paper, we propose a novel regularization term to train deep neural networks when using biased data.
The core idea of using mutual information is inspired by InfoGan~\cite{infogan}.
In constrast to the inspiring approach, we rather minimize the mutual information in order \textit{not} to learn.
By letting networks play minimax game, networks learn to categorize, while unlearning the bias.
The experimental results showed that the networks trained with the proposed regularization can extract bias-independent feature embedding, achieving the best performance in the most of the experiments.
Furthermore, our model performed better than ``Gray'' model which was trained with almost unbiased data, indicating the feature embedding becomes even more informative.
To conclude, we have demonstrated in this paper that the proposed regularization improves the performance of neural networks trained with biased data.
We expect this study to expand the usage of various data and to contribute to the field of feature disentanglement.

\section*{Acknowledgement}
This research was supported by Samsung Research.

\newpage
{\small
\bibliographystyle{ieee}
\bibliography{egbib}
}

\end{document}


\title{Learning Not to Learn: Training Deep Neural Networks with Biased Data \newline Supplementary Document}
\author{Byungju Kim$^1$ \qquad  Hyunwoo Kim$^2$ \qquad Kyungsu Kim$^3$ \qquad Sungjin Kim$^3$ \qquad Junmo Kim$^1$\\
School of Electric Engineering, KAIST, South Korea$^1$\\
Beijing Institute of Technology$^2$, \qquad Samsung Research$^3$\\
{\tt\small \{byungju.kim,junmo.kim\}@kaist.ac.kr}\\
{\tt\small \{hwkim\}@bit.edu.cn}, \qquad {\tt\small \{ks0326.kim,sj9373.kim\}@samsung.com}
}
\maketitle

\section*{Formulation of $\mathcal{L}_{MI}(\theta_f,\theta_h)$}
Here, we re-formulate regularization loss, $\mathcal{L}_{MI}(\theta_f,\theta_h)$, with more rigorous notations.
If it is not specifically mentioned, the numbering of equations indicates that of supplementary material (this document).
Our regularization loss is based on the mutual information between feature embedding and bias.
It can be expressed as follows:
\begin{equation}
\label{eq:mi_base}
\mathcal{I}(b(X);f(X)) = H(b(X)) - H(b(X) | f(X)).
\end{equation}
The Eq.~(\ref{eq:mi_base}) is the identical equation with Eq.~(4) in the paper.
The optimization problem to minimize the mutual information can be written as
\begin{equation}
\label{eq:argmin_1}
\begin{split}
\argmin_{\theta_f}\mathcal{I}(b(X);f(X)) & = \argmin_{\theta_f} H(b(X)) - H(b(X) | f(X)) \\
                   & = \argmin_{\theta_f} -H(b(X) | f(X)) \\
                   & = \argmin_{\theta_f} \mathbb{E}_{\tilde{F}\sim P_{f(X)}(\cdot)}[-H(b(X)|f(X)=\tilde{F})] \\
                   & = \argmin_{\theta_f} \mathbb{E}_{\tilde{x}\sim P_{X}(\cdot)}[-H(b(X)|f(X)=f(\tilde{x}))].
\end{split}
\end{equation}

By sampling $b(X)$, the conditional entropy in Eq.~(\ref{eq:argmin_1}) is reformulated as
\begin{equation}
\label{eq:cond_ent}
\begin{split}
-H(b(X)|f(X)=f(\tilde{x})) & = \mathbb{E}_{\tilde{b}\sim P_{b(X)|f(X)}(\cdot|f(\tilde{x}))}[\log P_{b(X)|f(X)}(\tilde{b}|f(\tilde{x}))] \\
& = \mathbb{E}_{\tilde{b}\sim Q_{b(X)|f(X)}(\cdot|f(\tilde{x}))}[\log Q_{b(X)|f(X)}(\tilde{b}|f(\tilde{x}))] \\
& \quad \quad s.t. \quad Q_{b(X)|f(X)}(b|f) = P_{b(X)|f(X)}(b|f)\textrm{ for all $b$ and $f$}.
\end{split}
\end{equation}

By substituting the Eq.~(\ref{eq:cond_ent}) for Eq.~(\ref{eq:argmin_1}), we have
\begin{equation}
\label{eq:argmin_2}
\begin{split}
\argmin_{\theta_f}\mathcal{I}(b(X);f(X)) & = \argmin_{\theta_f} \mathbb{E}_{\tilde{x}\sim P_{X}(\cdot)}[\mathbb{E}_{\tilde{b}\sim Q_{b(X)|f(X)}(\cdot|f(\tilde{x}))}[\log Q_{b(X)|f(X)}(\tilde{b}|f(\tilde{x}))]] \\
& \quad \quad s.t. \quad Q_{b(X)|f(X)}(b|f) = P_{b(X)|f(X)}(b|f)\textrm{ for all $b$ and $f$}.
\end{split}
\end{equation}

This is the identical with Eq.~(5) in the main paper.
Then, the $\mathcal{L}_{MI}$ can be reformulated as follows:

\begin{equation}
\label{eq:L_MI}
\begin{split}
\mathcal{L}_{MI}  = \ \  & \mathbb{E}_{\tilde{x}\sim P_{X}(\cdot)}[\mathbb{E}_{\tilde{b}\sim Q_{b(X)|f(X)}(\cdot|f(\tilde{x}))}[\log Q_{b(X)|f(X)}(\tilde{b}|f(\tilde{x}))]] \\
& + \mu D_{KL}(P_{b(X)|f(X)} || Q_{b(X)|f(X)}).
\end{split}
\end{equation}

Now, we parametrize the $Q_{b(X)|f(X)}(b|f)$ as $h(b|f;\theta_{h})$, and try to make $h(b|f;\theta_{h})$ close to $P_{b(X)|f(X)}(b|f)$ by supervised learning.
Therefore, the $\mathcal{L}_{MI}(\theta_f,\theta_h)$ is
\begin{equation}
\label{eq:L_MI_theta}
\begin{split}
\mathcal{L}_{MI}(\theta_f,\theta_h)  = \ \  & \mathbb{E}_{\tilde{x}\sim P_{X}(\cdot)}[\mathbb{E}_{\tilde{b}\sim h(\cdot|f(\tilde{x}))}[\log h(\tilde{b}|f(\tilde{x}))] \\
& + \mu \mathcal{L}_{c}(b(\tilde{x}),h(f(\tilde{x})))],
\end{split}
\end{equation}
where $b(\tilde{x})$ is a one-hot label vector for bias in image $\tilde{x}$.


\section*{Cleaning Procedure for IMDB Face Dataset}
To clean the noisy label from the IMDB Face dataset, we used networks pretrained with Adience dataset.
We estimated the age and gender for all the individuals in the IMDB dataset.
We discarded an image of the estimations that do not match with the ground truth labels.
Unlike IMDB dataset, the age label of Adience dataset is provided in terms of interval:
(0-2), (4-6), (8-13), (15-20), (25-32), (38-43), (48-53), and (60-).
To determine whether the age estimation matches the ground truth label or not, we set a maring $M$.
With estimated age (L, H) for an image, we consider the age estimation matches for age label of the image if the estimation is within range of (L-M, H+M).

\section*{Color Bias in MNIST Dataset}
To intentionally plant color bias to MNIST dataset, every image is colored.
Ten colors were selected and each color is assigned to a digit category.
The colors and corresponding categories are enumerated in Table~\ref{table:colors}.
\begin{table}[h]
\centering      
\begin{tabular}{c|l|c|c|c}
Digit & \multicolumn{1}{c|}{Color Name} &  Mean Color   & Sampled Mean - Train &  Sampled Mean - Test \\ \hline
0     & Crimson                         & (220, 20, 60) & (214, 39, 76)        & (148,134,116)        \\
1     & Teal                            & (  0,128,128) & ( 29,127,127)        & (151,136,116)        \\
2     & Lemon                           & (253,233, 16) & (225,211, 40)        & (147,127,116)        \\
3     & Bondi Blue                      & (  0,149,182) & ( 29,129,194)        & (152,133,115)        \\
4     & Carrot orange                   & (237,145, 33) & (221,128, 54)        & (147,132,115)        \\
5     & Strong Violet                   & (145, 30,188) & (143, 43,184)        & (152,134,112)        \\
6     & Cyan                            & ( 70,240,240) & ( 72,219,219)        & (148,129,118)        \\
7     & Your pink                       & (250,197,187) & (223,186,186)        & (148,134,120)        \\
8     & Lime                            & (210,245, 60) & (201,221, 63)        & (151,133,115)        \\
9     & Maroon                          & (128,  0,  0) & (127, 28, 28)        & (145,133,116)        \\ \hline
\end{tabular}
\caption{Mean colors for sampling and sampled mean colors. Although the colors are presented with integers in [0, 255], they were normalized into [0, 1]. The sampled mean colors of test images show that the test set is independent of the color bias unlike the training set}
\label{table:colors}
\end{table}

Originally, 20 distinct colors are proposed.\footnote{https://sashat.me/2017/01/11/list-of-20-simple-distinct-colors/}
We have selected ten colors with minor modification.
With the mean colors, a color for each image is sampled following Algorithm~\ref{algo:sample}.
\begin{algorithm}[h]
\caption{Sample a color for each image.}
\begin{algorithmic} 
\REQUIRE mean color ($r^m$, $g^m$, $b^m$) $\in \mathbb{R}^3$, variance $\sigma^2$
\STATE Initialize sampled color $\mathcal{C}$ = []
\FOR{$c^m$ in [$r^m$, $g^m$, $b^m$]}
\WHILE{True}
\STATE Sample a color $c\sim\mathcal{N}(c^m, \sigma^2$)
\IF{$0<c<1$}
\STATE Append $c$ to $\mathcal{C}$
\STATE \bf{break}
\ENDIF
\ENDWHILE
\ENDFOR
\end{algorithmic}
\label{algo:sample}
\end{algorithm}

Training images were colored depending on their digit labels.
For example, if an image to colorize is from category zero, the color is sampled with mean color (220, 20, 60).
Therefore, images from the same digit category are colorized with similar colors.
In contrast, colors of the test images should be independent of their categories.
To this end, a mean color is sampled uniformly in advance to the algorithm~\ref{algo:sample}.
As a result of sampling, the mean colors of each category are presented in Table~\ref{table:colors}.